\documentclass[letterpaper, 10 pt, conference]{ieeeconf}  % Comment this line out if you need a4paper

\IEEEoverridecommandlockouts                              % This command is only needed if 
\usepackage{mathtools, bm}

\usepackage{graphics} % for pdf, bitmapped graphics files
\usepackage{epsfig} % for postscript graphics files
\usepackage{amsmath} % assumes amsmath package installed
\usepackage{amssymb}  % assumes amsmath package installed
\usepackage{amsfonts}
\usepackage{subfiles}
\graphicspath{{images/}{../images/}}
\usepackage{csquotes}
\usepackage{gensymb}
\usepackage{pifont}
\usepackage{subcaption}
\usepackage{textcomp}
\usepackage[table]{xcolor}
\usepackage[english]{babel}
\usepackage{biblatex}
\usepackage[export]{adjustbox}
\usepackage{pdfpages}
\usepackage[font={footnotesize}]{caption}
\usepackage{blindtext}
\usepackage{soul}
\usepackage{siunitx}
\usepackage{hyperref}

\title{Redundant Perception and State Estimation for Reliable Autonomous Racing\vspace{-10pt}}

\author{Nikhil Gosala$^{*1}$, Andreas B\"uhler$^{*1}$,  Manish Prajapat$^{*1}$, Claas Ehmke$^{*1}$, Mehak Gupta$^{*2}$, Ramya Sivanesan$^{*2}$\\  Abel Gawel$^{1}$, Mark Pfeiffer$^{1}$, Mathias B\"urki$^{1}$, Inkyu Sa$^{1}$, Renaud Dub\'e$^{1}$, and Roland Siegwart$^{1}$% 
\thanks{* The authors contributed equally to this work.}%
\thanks{$^{1}$ Authors are with the Autonomous Systems Lab, ETH Z\"urich, Z\"urich.}%
\thanks{$^{2}$ Authors are with the CVG Group, ETH Z\"urich, Z\"urich.}%
}

\begin{document}

\maketitle
\thispagestyle{empty}
\pagestyle{empty}

\begin{abstract}
In autonomous racing, vehicles operate close to the limits of handling and a sensor failure can have critical consequences.
To limit the impact of such failures, this paper presents the redundant perception and state estimation approaches developed for an autonomous race car. 
Redundancy in perception is achieved by estimating the color and position of the track delimiting objects using two sensor modalities independently. 
Specifically, learning-based approaches are used to generate color and pose estimates, from LiDAR and camera data respectively.
The redundant perception inputs are fused by a particle filter based SLAM algorithm that operates in real-time.
Velocity is estimated using slip dynamics, with reliability being ensured through a probabilistic failure detection algorithm.
The sub-modules are extensively evaluated in real-world racing conditions using the autonomous race car \textit{gotthard driverless}, achieving lateral accelerations up to 1.7G and a top speed of 90km/h.

\end{abstract}

\section{Introduction}
\label{sec:introduction}
	
Autonomous driving and its racing counterpart have received a lot of attention since the inception of the DARPA challenge in 2004~\protect\cite{cit:Singh09}.
Fuelled by racing series like Roborace and Self Racing Cars, state-of-the-art algorithms have been developed to fulfill the requirements of real-world racing conditions~\protect\cite{cit:audi-tt-control}~\protect\cite{cit:alex-liniger-control}. 
Despite major technological advances, developing reliable autonomous vehicles remains a challenge. 
For instance, in 2016, an autonomous vehicle failure was reported once every three hours in California alone~\protect\cite{cit:ieee-autonomous-failures}. 
To make autonomous vehicles safe and reliable, the robustness of both the sensor setup and the algorithms has to be enhanced.
This paper aims to improve the reliability of the perception and state estimation pipelines by introducing algorithms that provide redundancy for processing data generated by multiple complementary sensors.

Several works in multi-modal perception focus on fusing measurements from different sensors to accurately estimate the robot state and map the environment.
For example, multi-sensor fusion approaches are used to enhance object recognition and tracking~\protect\cite{cho2014multi,premebida14pedestrian} or to find regions suitable for driving~\protect\cite{li2014drivable}. 
Although these approaches improve robustness and accuracy by fusing sensors, they do not provide redundancy in case of a sensor failure. For instance, if a visual sensor fails, the perception pipeline could lose either depth or semantic information, potentially reducing the robustness of the overall system. 
Sensor failure detection is also an active research area with steps being made towards outlier rejection~\protect\cite{cit:Iglesia18}, and detecting sensor malfunctions due to system attacks~\protect\cite{bezzo14attack}. 
So far, redundancy has been achieved by replicating the sensor setup and running the pipelines using a voting based approach~\protect\cite{cit:underground-mining, cit:odin}.
However such approaches result in higher costs and computational demands. 

This paper presents a redundant architecture that enables reliable Simultaneous Localization and Mapping (SLAM) for an autonomous race car. 
The reliability of our perception module is improved by estimating the color and position of landmarks demarcating the track using both LiDAR and camera independently.  
Furthermore, the reliability of the velocity estimate is enhanced by the use of a failure detection module that can detect and isolate faulty sensors.
These functionalities thus allow us to safely operate an autonomous race vehicle even under single-sensor failures.
The approaches are experimentally evaluated with \textit{gotthard driverless}, the autonomous race car that went on to win multiple Formula Student Driverless competitions in 2018.
 
The main contributions of this paper are (i) a learning-based approach to estimate landmark colors using LiDAR measurements, (ii) an efficient stereo-matching algorithm that reduces search space using a learning-based approach, (iii) an EKF-based slip aware velocity estimator with probabilistic failure detection, and (iv) a particle filter based SLAM algorithm fusing multi-modal landmark observations.

A video presenting the perception and state-estimation pipelines along with the real-world performance of \textit{gotthard driverless} can be found online\footnote{\url{www.youtube.com/watch?v=ir_uqEYuT84}}.

\begin{figure}
    \centering
    \includegraphics[width=0.73\linewidth]{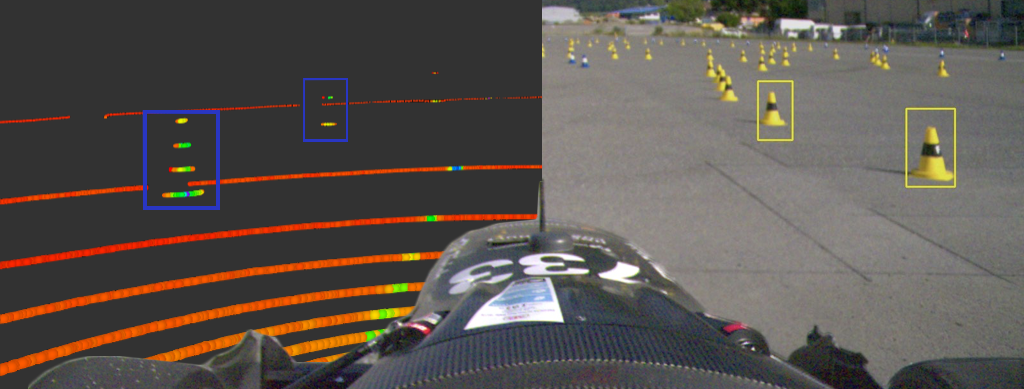}
    \setlength{\abovecaptionskip}{3pt}
    \setlength{\belowcaptionskip}{-20pt}
    \caption{The world as viewed by the event winning autonomous race car \textit{gotthard driverless} using LiDAR (left) and cameras (right).}
    \label{fig:gotthard-driverless}
\end{figure}

\section{Problem Statement}
\label{sec:problem-statement}
    
The objective of this work is to enable a race car to autonomously complete multiple laps of an unknown race track ($\sim$\SI{500}{\meter} long) without any human intervention and in a single attempt. 
The left and right boundaries of the race track are assumed to be demarcated using blue and yellow cones respectively~\protect\cite{cit:fsg-rules}.
The two main challenges faced in such scenarios are (i) the lack of prior knowledge of the race track, and (ii) the possibility of a sensor failure hindering the operation of the car.

To enable autonomous navigation, the race car is equipped with a 3D \ LiDAR, and three color cameras in a mono and stereo setup.
An inertial navigation system (INS), an optical ground speed sensor (GSS), and four wheel speed sensors (WSS) allow for real-time state estimation.

\section{Method}
\label{sec:concept}
	
This section describes the approaches developed for the reliable operation of an autonomous race car.
To ensure that the car stays within the track limits, knowledge of the cones' position and color is required.
The observations gathered by the perception sensors are fused with the velocity estimate to guarantee reliable mapping and state propagation, thus ensuring successful navigation around the race track.

\subsection{LiDAR Cone Detection and Color Estimation}
\label{subsec:lidar}
	
\begin{figure}
	\centering
  \begin{subfigure}[b]{\columnwidth}
  	\centering
   	\includegraphics[width=0.7\linewidth]{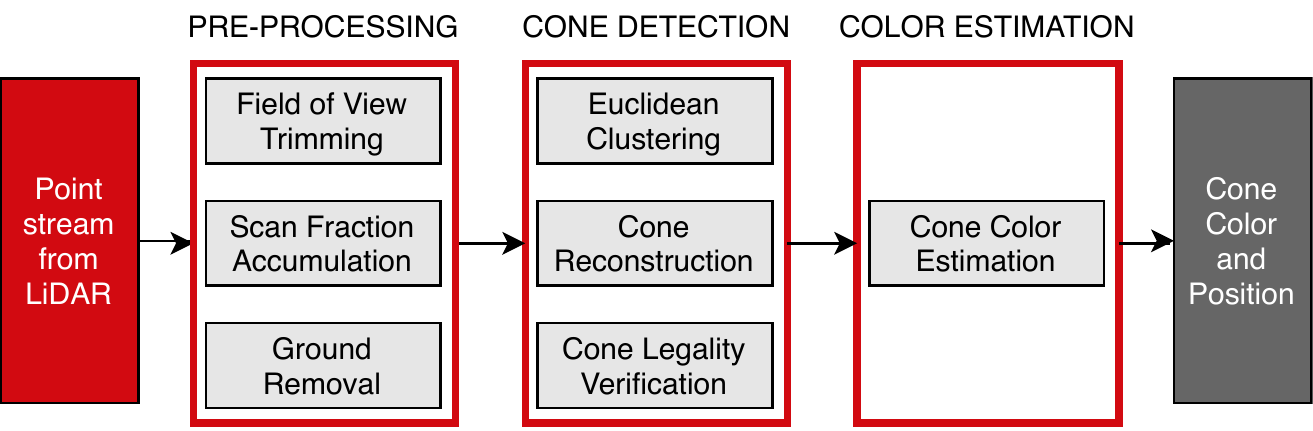}
    \setlength{\abovecaptionskip}{2pt}
    \caption{}
    \label{fig:lidar-pipeline-diagram}
  \end{subfigure}
  \begin{subfigure}[b]{0.35\columnwidth}
   	\includegraphics[width=\linewidth,left]{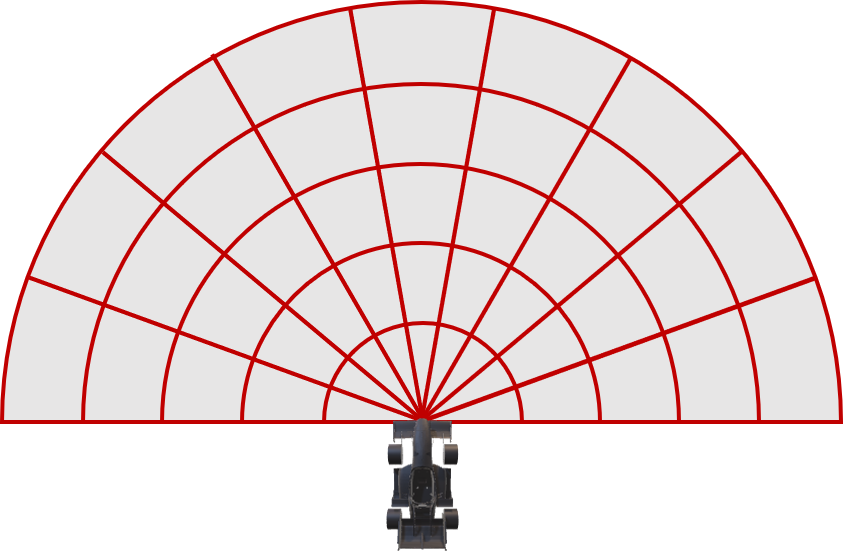}  
    \setlength{\abovecaptionskip}{-10pt}
    \caption{}
    \label{fig:lidar-adaptive-ground-removal-segmentation}
  \end{subfigure}
  \begin{subfigure}[b]{0.35\columnwidth}
   	\includegraphics[width=\linewidth, right]{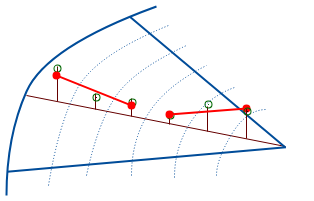} 
    \setlength{\abovecaptionskip}{-10pt}
    \caption{}
    \label{fig:lidar-adaptive-ground-removal-side}
  \end{subfigure}
  \setlength{\abovecaptionskip}{1pt}
  \setlength{\belowcaptionskip}{-20pt}
  \caption{(a) The LiDAR pipeline for cone color and position estimation in real-time. (b) Top view of the segmentation of the ground into sectors and bins. (c) Isometric view of the adaptive ground removal in one sector. Red lines represent the fitted ground lines \protect\cite{cit:adaptive-ground-removal}.}
\end{figure}

The first step towards redundancy in perception is achieved by estimating the color and position of cones using LiDAR only. 
The sub-system architecture is depicted in Figure \ref{fig:lidar-pipeline-diagram}, the main elements of which are described below.

\subsubsection{Pre-Processing}
\label{subsubsec:lidar-pre-processing}
Motion distortion in LiDAR scans is compensated using the velocity estimates, after which distortions as large as \SI{2}{\meter} in a single scan are reduced to only \SI{2.6}{\cm}.
The ground points from the resulting point cloud are removed using an adaptive ground removal algorithm~\protect\cite{cit:adaptive-ground-removal} that adapts to changes in inclination of the ground. 
The ground is split into multiple sectors and bins (Figure~\ref{fig:lidar-adaptive-ground-removal-segmentation}) and lines are fit through the lowest points of each bin (Figure~\ref{fig:lidar-adaptive-ground-removal-side}). Finally, all points within a threshold of the closest line are removed.

\subsubsection{Cone Reconstruction and Filtration}
\label{subsubsec:lidar-cone-reconstruction-and-filtration}
Ground removal also results in the removal of nearly $64\%$ of cone points on average, which reduces the number of points per cone and makes cone identification challenging. 
This is overcome by clustering the points after ground removal using the euclidean distance based approach, and reconstructing a small cylindrical area around each cluster using points from the distortion-free point cloud. 
The reconstructed clusters are passed through a filter that checks whether the number of points in the cluster matches the expected number of points in a cone at that distance, which is computed using \eqref{eqn:cone-filter}:
\begin{equation}
\label{eqn:cone-filter}
E(n_{d}) = \frac{1}{2} \times \frac{h_{c}}{2*d*\tan(\frac{r_v}{2})} \times \frac{w_{c}}{2*d*\tan(\frac{r_h}{2})}
\end{equation}
where $n_{d}$ is the number of points at distance d, $h_{c}$ and $w_{c}$ are the height and width of the cone respectively, and $r_v$ and $r_h$ are the vertical and horizontal angular resolutions of the LiDAR respectively. 
The clusters that pass through the filter are then propagated to the color estimation module.

\subsubsection{Color Estimation}
\label{subsubsec:lidar-color-estimation}
\begin{figure}
	\centering
  \begin{subfigure}[b]{0.8\columnwidth}
  	\centering
   	\includegraphics[width=\linewidth]{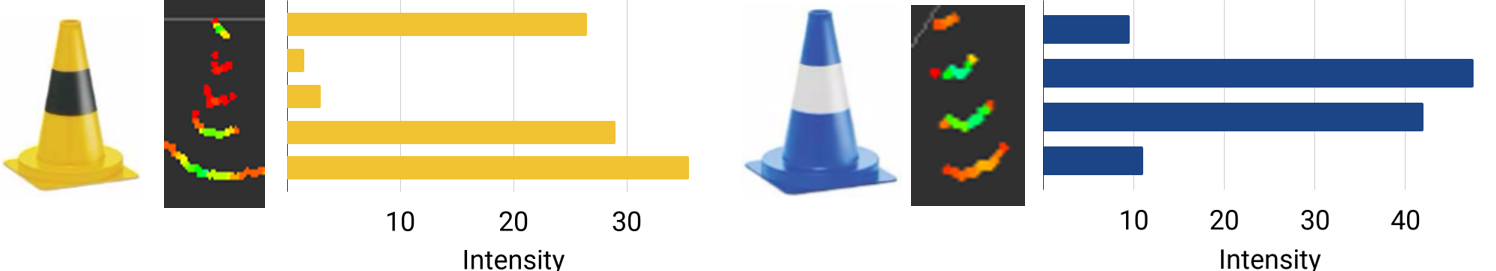}  
   	\setlength{\abovecaptionskip}{-11pt}
    \caption{}
    \label{fig:lidar-cone-color-intensity-gradient}
  \end{subfigure}
  \begin{subfigure}[b]{0.9\columnwidth}
  	\centering
   	\includegraphics[width=\linewidth]{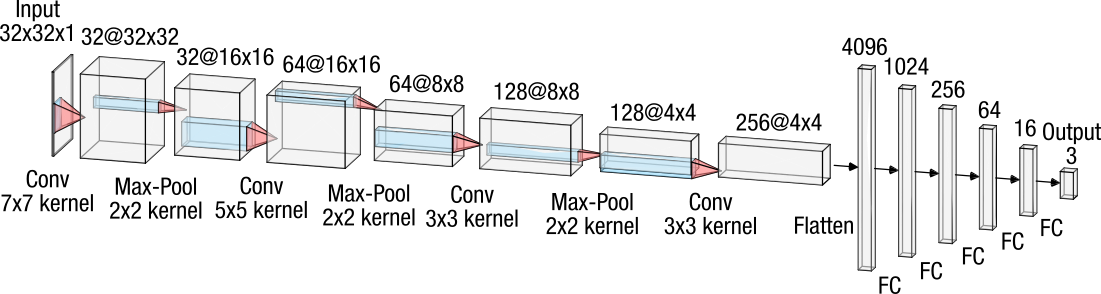}
   	\setlength{\abovecaptionskip}{-14pt}
    \caption{}
    \label{fig:lidar-color-cnn}
  \end{subfigure}
  \setlength{\abovecaptionskip}{0pt}
  \setlength{\belowcaptionskip}{-7pt}
  \caption{(a) Intensity gradient for yellow and blue cone. (b) CNN used for color estimation.}
\end{figure}

The color estimation uses the repeatable intensity patterns in the point cloud obtained from the cones.
Figure \ref{fig:lidar-cone-color-intensity-gradient} shows the cones and the varying intensity order as one moves along the vertical axis of the cones. 
This differing intensity order is capitalized upon, and the color is estimated using a Convolutional Neural Network (CNN) (see Figure \ref{fig:lidar-color-cnn}). 
To improve the generalization of the network, dropout and batch normalization are used.
Additionally, incorrect predictions are penalized using an asymmetric cross-entropy loss function that penalizes misclassifications by a factor of 100.
The CNN accepts a 32x32 grayscale image of a cone with pixel values representing intensities of points in the point cloud, and outputs the probability of each cone being \textit{blue}, \textit{yellow}, and \textit{unknown}.

We hypothesize that compared to a rule-based classification approach, the CNN offers higher robustness to noise and is capable of learning hidden patterns from the input data.
Furthermore, the color estimation is capped at \SI{5}{\meter}, because the sparsity of the point cloud above this distance does not allow for a distinction between the color patterns.

\subsection{Visual Cone Detection and Stereo Pose Estimation}
\label{subsec:vision}
	
\begin{figure}
   \centering
   \includegraphics[width=0.8\linewidth]{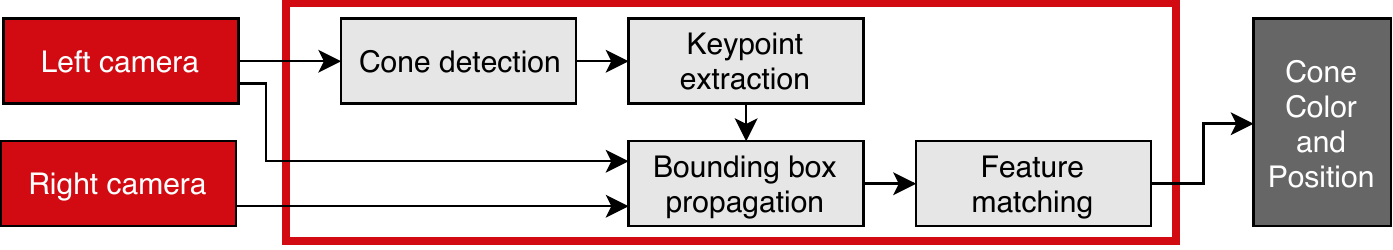}
   \setlength{\abovecaptionskip}{4pt}
   \setlength{\belowcaptionskip}{-20pt}
   \caption{Vision system architecture with images as the input, and color and 3D position estimates of the cones as output.}
   \label{fig:vision-stereo}
\end{figure}

The cones' colors and positions are estimated by the stereo camera in addition to the LiDAR. The presence of multiple identical cones in an image poses a challenge for matching corresponding cones across images. This is overcome efficiently by detecting cones using YOLOv2 \protect\cite{cit:yolov2} in only one image, and spatially propagating the bounding boxes to the other by exploiting the prior knowledge of their appearance. The 3D position estimate is then improved by triangulating only the specific patches of interest instead of the complete stereo image pair. The major components of the pipeline are described below and its architecture is shown in Figure \ref{fig:vision-stereo}.

\subsubsection{Cone Detection}
\label{subsubsec:vision-cone-detection}
YOLOv2, which offers good accuracy while being computationally efficient is used to detect cones in the images (Figure \ref{fig:vision-propagation-left}). It is trained on three classes: \textit{blue}, \textit{yellow}, and \textit{orange} cones. The network parameters like the anchor box size, non-maximum suppression and confidence thresholds are tuned on a self-acquired dataset to reduce both false positives and inaccurate bounding boxes. The network is trained on images with varying illumination and weather conditions for better generalization and robustness in real-world applications.

\subsubsection{Keypoint Extraction}
\label{subsubsec:vision-keypoint-regression}
As the shape and size of the cones demarcating the track are standardized and fixed, a representative 3D model of a cone is generated and is used along with the corresponding 2D keypoints to calculate the pose of the cone with respect to the camera via the perspective-n-point algorithm (PnP) \protect\cite{cit:pnp-vision}. To extract these 2D keypoints in a bounding box, a neural network\footnote{The neural network architecture is part of a separate contribution.} is developed \protect\cite{cit:kp-ankit}. Inspired by classical computer vision, where corners are among the most prominent features in images, this neural network regresses on 7 such keypoints (Figure \ref{fig:vision-keypoint}) using cross-ratio in its loss function. This results in 3D position estimates of cones in the left camera's coordinate system.

\subsubsection{Spatial Bounding Box Propagation}
\label{subsubsec:vision-spatial-bounding-box-propagation}
The 3D position estimates of the cones obtained from the left camera's image are expressed in the right camera's coordinate system using the stereo camera calibration. The cones are then projected onto the right camera's image plane. Using stereo geometry, the bounding boxes from the left camera's image are thus spatially propagated to the right camera's image. The precision of the propagation is enhanced by introducing constraints based on disparity and epipolar geometry.

\subsubsection{Feature Matching and Triangulation}
\label{subsubsec:vision-feature-matching-and-triangulation}
\begin{figure}
	\centering
  \begin{subfigure}[b]{0.38\columnwidth}
  	\centering
   	\includegraphics[width=\linewidth]{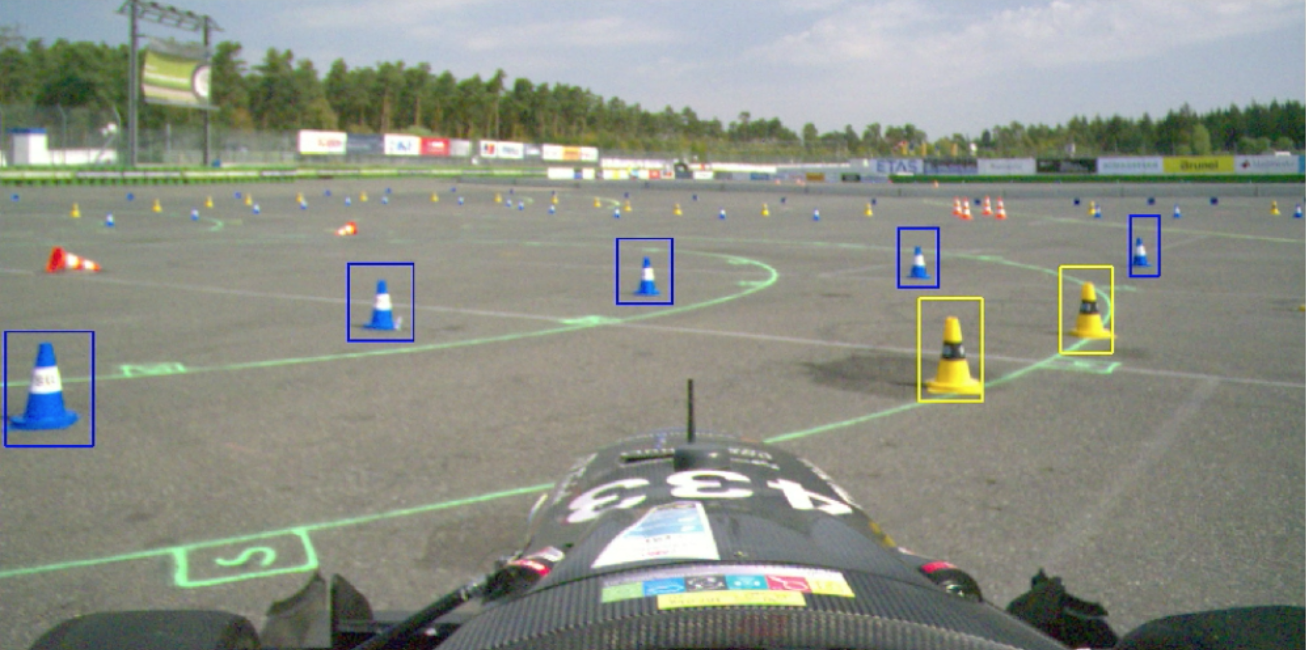} 
   	\setlength{\abovecaptionskip}{-11pt}
    \caption{}
    \label{fig:vision-propagation-left}
  \end{subfigure}
  \begin{subfigure}[b]{0.29\columnwidth}
  	\centering
   	\includegraphics[width=\linewidth]{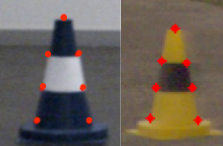}  
   	\setlength{\abovecaptionskip}{-11pt}
    \caption{}
    \label{fig:vision-keypoint}
  \end{subfigure}
  \begin{subfigure}[b]{0.25\columnwidth}
  	\centering
   	\includegraphics[width=\linewidth]{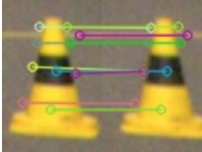} 
   	\setlength{\abovecaptionskip}{-11pt} 
    \caption{}
    \label{fig:vision-feature-matching-yellow}
  \end{subfigure}
  \setlength{\abovecaptionskip}{0pt}
  \setlength{\belowcaptionskip}{-20pt}
  \caption{(a) Bounding boxes from YOLOv2 on image from left camera. (b) Keypoints regressed on cones (c) Feature matching in corresponding bounding boxes.}
\end{figure}

The final step in obtaining the improved 3D position estimate is to triangulate the corresponding pair of left and right bounding boxes. SIFT features \protect\cite{cit:sift-vision} (robust to scale, rotation and illumination) are extracted and matched across the bounding box pair using brute-force matching (Figure~\ref{fig:vision-feature-matching-yellow}). The matched features are triangulated and the median of these triangulated points is then used as the 3D position estimate of the cone. Median is preferred to mean because the latter is susceptible to outlier matches which would result in incorrect 3D position estimates.

\subsection{Velocity Estimation}
\label{subsec:velocity_estimation}
	
Robust and accurate velocity estimates are key for reliable operation of autonomous cars. 
They are used to propagate poses, compensate for motion distortion in LiDAR data, and influence accuracy of SLAM and efficiency of control algorithms.
Compared to typical mobile ground robots, race cars have high wheel slip, up to 20\% for optimal longitudinal acceleration \protect\cite{cit:Pacejka2012}, which strongly biases wheel odometry. 
In addition, velocity sensors like GNSS and GSS are prone to failure or bias in rough environment (e.g. cloudy, cluttered environments~\protect\cite{cit:AD_NAV}, or wet surfaces~\protect\cite{cit:Kistler}). These challenges are addressed by estimating the slip within a probabilistic framework. 

There exists a variety of filtering~\protect\cite{cit:filters} and batch optimization approaches for state estimation~\protect\cite{cit:Mehmet2014}. EKF was selected due to its computational efficiency, ease of debugging, access to state covariances and due to its proven performance in real-world applications~\protect\cite{cit:Iglesia18}.

\begin{figure}
	\centering
    \includegraphics[width=0.4\textwidth, angle=0]{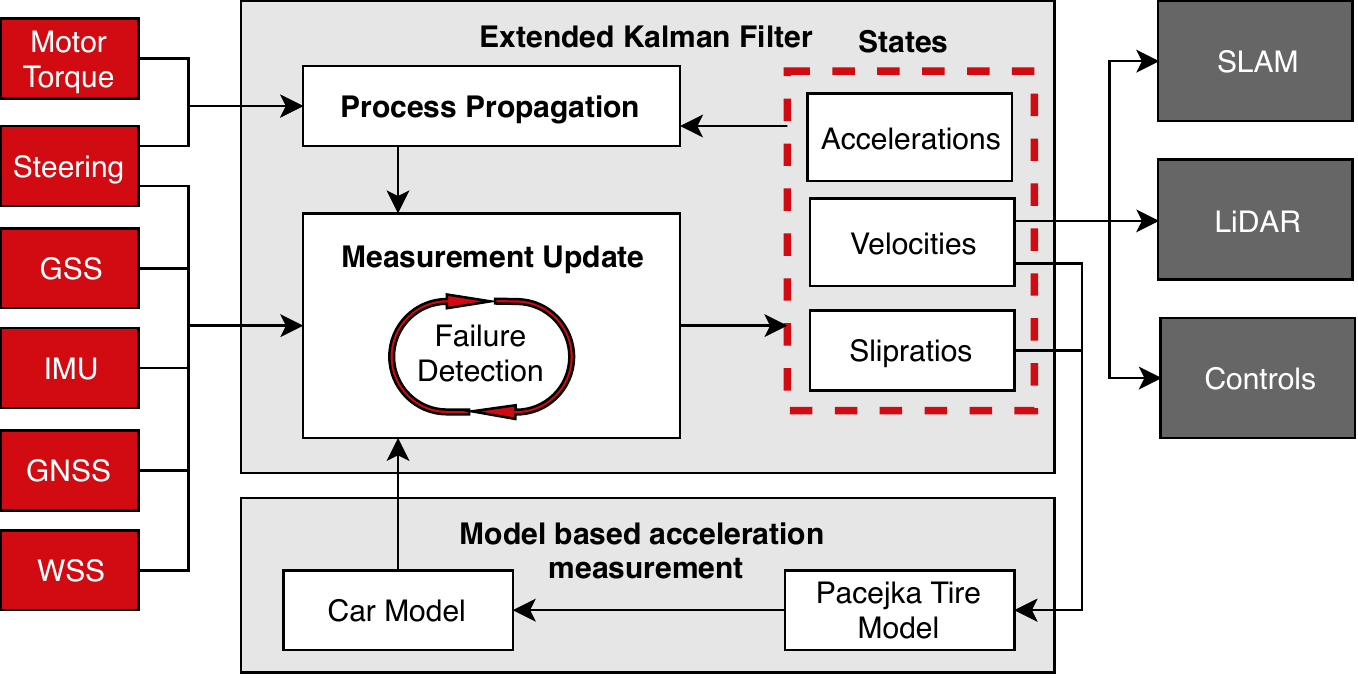}
    \setlength{\abovecaptionskip}{4pt}
    \setlength{\belowcaptionskip}{-20pt}
    \caption{A simplified velocity estimation architecture}
    \label{fig:VE_architecture}
\end{figure}

This section presents the process and measurement models followed by a failure detection module that isolates the posterior from false likelihood measurements. A simplified architecture is shown in figure~\ref{fig:VE_architecture}.

\subsubsection{Process Model}
\label{process_model}
The car is assumed to remain in contact with a flat surface, allowing the model to be built in 2D. The state vector $\mathbf{x}\in\mathbb{R}^{9\times1}$ is defined as
\begin{comment}
\begin{align}\label{eq:state}
\begin{split}
\mathbf{x} &= [\mathbf{v}^T, r, \mathbf{a}^T, \mathbf{sr}^T]^{T} \\
where, \quad
\mathbf{v} &= [v_x, v_y]^T \\
\mathbf{a} &= [a_x, a_y]^T \\
\mathbf{sr} &= [sr_{fl}, sr_{fr}, sr_{rl}, sr_{rr}]^T
\end{split}
\end{align}
\end{comment}
\begin{align}\label{eq:state}
\begin{split}
\mathbf{x} &= [\mathbf{v},\;r,\;\mathbf{a},\;\mathbf{sr}]^{T}, \\
\end{split}
\end{align}
where $\mathbf{v}=[v_{\stxt{x}}, v_{\stxt{y}}]$ and $r$ represent the linear and angular velocities respectively, $\mathbf{a}=[a_{\stxt{x}}, a_{\stxt{y}}]$ denotes the linear accelerations (note that we omit the vertical component of velocity and acceleration), and $\mathbf{sr_{\stxt{ij}}}=[sr_{\stxt{FL}}, sr_{\stxt{FR}}, sr_{\stxt{RL}}, sr_{\stxt{RR}}]$ denotes the slip ratio of wheel ij, where i $\in \{\text{Front} ,\; \text{Rear} \}$ and j $\in \{\text{Left} ,\; \text{Right} \}$, defined as:
\begin{equation}\label{eq:slipratio_equation}
sr_{{\stxt{ij}}} = \left\{
	\begin{array}{l l}
	\frac{\omega_{{\stxt{ij}}} \cdot R - V_{{\stxt{ij}}} }{V_{{\stxt{ij}}}} & \quad \text{if $|V_{ij}|>0$}\\
	\quad 0 & \quad \text{if $|V_{{\stxt{ij}}}|=0$,}
	\end{array} \right. 
\end{equation}
where $\omega_{{\stxt{ij}}}$ and $v_{{\stxt{ij}}}$ are the angular and linear velocities of wheel ij and $R$ is the radius of the wheel. 
The process model represents a prior distribution over the state vector wherein the velocity is propagated using a constant acceleration model and slip ratios are propagated using the dynamics derived by time differentiation of slip ratio given by \ref{eq:slipratio_equation}. The process model is defined as:
\begin{align}
\begin{split}
\dot{\mathbf{v}} &= \mathbf{a} + \left[ v_{\stxt{y}} r, \, -v_{\stxt{x}} r \right]^T+ \mathbf{n_{\stxt{v}}}\\
\dot{r} &= f_{\stxt{M}}(\mathbf{sr}, \mathbf{v}, r, \delta) + n_{\stxt{r}} \\
\dot{\mathbf{a}} &= \mathbf{n_{\stxt{a}}} \\
\dot{\mathbf{sr}} &= \frac{\mathbf{T_{\stxt{M}}} \cdot R}{\mathbf{V_{\omega x}} \cdot I_{\omega}} + \frac{\mathbf{sr}}{\mathbf{V_{\omega x}}} \cdot (\frac{C_\sigma \cdot R}{I_{\omega}} - a_{{\stxt{x}}}) - \frac{a_{\stxt{x}}}{\mathbf{V_{\omega x}}} + \mathbf{n_{\stxt{sr}}}\\
\end{split}
\label{eq:process_model}
\end{align}
where $f_{\stxt{M}}(\cdot)$ computes yaw moment based on tire forces estimated using a linear function of longitudinal and lateral slip (see Chapter I \protect\cite{cit:Moustapha2013}). Motor torque $\mathbf{T_{\stxt{M}}} \in \mathbb{R}^{4\times1}$ and steering angle $\delta$ are input to the process model and measured using a current sensor and an encoder respectively. $C_{\sigma}$ is the longitudinal tire stiffness~\protect\cite{cit:Moustapha2013}, $\mathbf{V_{\omega x}} \in \mathbb{R}^{4\times1}$ is the longitudinal velocity of the wheel hub, $I_{\omega}$ is the moment of inertia of the wheel, and $\mathbf{n}_{\{\cdot\}}$ is the i.i.d. Gaussian white noise. These terms enable the process model to capture the fact that the probability of slippage increases with increase in motor torque.

\subsubsection{Measurement Model}
\label{measurement_model}
The measurements from all the sensors can be combined to update one or more of the following state variables:
\subsubsection*{Slip Ratio} Slip ratios are updated using the WSS measurements. Observability analysis~\protect\cite{cit:Iglesia18} concludes that with a certain combination of faulty sensors, the state variable becomes unobservable. To predict velocities in such cases, the model is switched from a full dynamic model to its partial kinematic counterpart by updating the slip ratios with a \emph{zero slip ratio update} (ZSU). High changes in wheel speeds are captured as slip by the process model and even tough the ZSU later shrinks the slip ratio in the update step of the EKF, the velocity estimate stays reliable.

\subsubsection*{Acceleration} Accelerations are estimated using the IMU and the car's dynamic model.
The IMU is considered reliable, yet on inclined surfaces the 2D assumption is violated and the gravity vector corrupts the true lateral and longitudinal accelerations. On the other hand, the dynamic model for calculating acceleration is based on the Pacejka model~\protect\cite{cit:Pacejka2012}, which is accurate for small slip but influenced by environmental conditions. Robustness is increased by fusing from these two sources, which is possible by having accelerations as a part of the state vector.

\subsubsection*{Velocity} Linear velocities are updated using the GSS, GNSS, and WSS whereas the angular velocity is observed using the IMU in addition to the above three.

The measurement model $\mathbf{z}$ $\in$ $\mathbb{R}^{13\times1}$ is given by:
\begin{align}\label{eq:sensor_model}
\begin{split}
\mathbf{z_{\stxt{v}}} &= h_{\stxt{v}}(\mathbf{x}) =   \mathbf{R}(\theta_{{\stxt{s}}}) (\mathbf{v} + [-r\, \mathbf{p}_{\stxt{s, y}}, \, r\, \mathbf{p}_{\stxt{s, x}}]^T) + \mathbf{{n}_{z_{v}}}\\
z_r &= h_{r}(\mathbf{x}) = r + n_{z_{r}}\\
\mathbf{z_{\mathbf{a}}} &= h_{a}(\mathbf{x}) = \mathbf{a} + \mathbf{n_{z_a}} \\
\mathbf{z_{\omega}} &= h_{\mathbf{\omega}}(\mathbf{x}) = \mathbf{V_{\omega x}} \cdot (\mathbf{sr}+1) / R + \mathbf{n_{z_{\omega}}} \\
\end{split}
\end{align}
Here, $\mathbf{R}(\theta_{\stxt{s}})$ denotes the rotation matrix where $\theta_{\stxt{s}}$ is the orientation of the sensor in the car frame and $n_{\{\cdot\}}$ is the i.i.d. Gaussian noise that corrupts the sensor measurements.

\subsubsection{Sensor Failure Detection}
\label{failure_detection}
Since inaccurate sensor measurements (e.g. sensor failure) can result in poor state estimation for the next iteration, recognition of such abnormal sensor status is critical for the recursive algorithm. The sensor faults can be classified as \textit{outlier} (e.g. spikes in measurements), \textit{drift} and \textit{null}. A Chi-square-based approach similar to \protect\cite{cit:Iglesia18} for outlier detection, and a variance based sensor isolation for drift detection given by \ref{eq:var_test} are implemented.
\begin{equation}\label{eq:var_test}
\vspace{-2pt}
\sum_{i=1}^{n} (\mathbf{z_{\stxt{i}}}-{\mu}_{\mathbf{\stxt{z}}})^2  < k
\end{equation}
$\mu_{\mathbf{\stxt{z}}}$ represents the mean of the sensor measurement. For each measurement, the variance is calculated with $n(>2)$ number of sensors used to measure that state variable. If the variance exceeds a tunable parameter $k$, sensors are removed progressively until the sensor with the highest contribution to the variance is rejected at the given time instance.

\subsection{Localization and Mapping}
	
\begin{figure}
    \centering
    \includegraphics[width=0.95\linewidth]{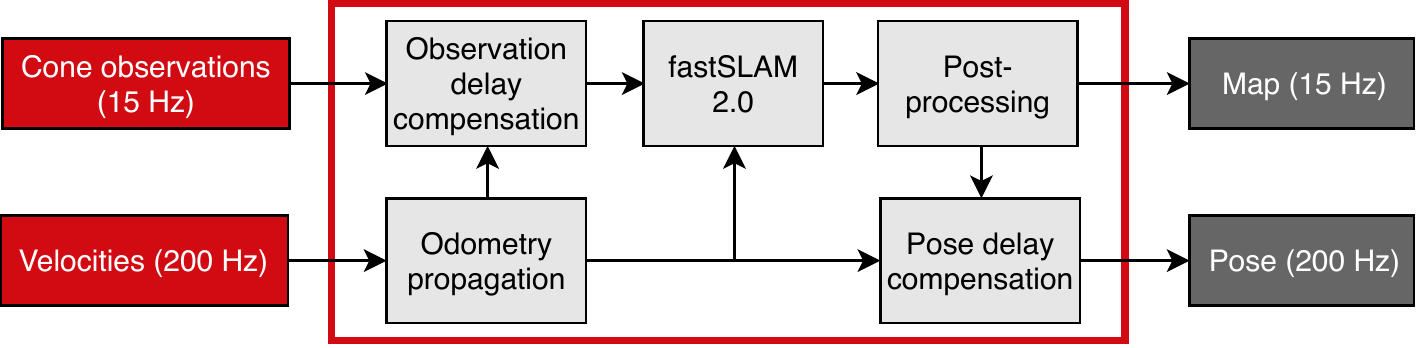}
    \setlength{\abovecaptionskip}{4pt}
    \setlength{\belowcaptionskip}{-20pt}
    \caption{SLAM Architecture during mapping phase with observations and velocity estimate as inputs and a resulting map and pose estimate as output. }
    \label{fig:slam_arch}
\end{figure}

To unfold the car's full dynamic potential, a planning horizon with at least \SI{2}{\second} look-ahead is required.
At high speeds it is infeasible to perceive upcoming corners for such a long horizon ($\sim$\SI{30}{\meter}).
Therefore, a map is built at low speeds and afterwards used to localize the car and plan manoeuvres.
Since the landmarks (LMs) are of similar appearance and can only be distinguished by their color and position, the algorithm should be able to incorporate probabilistic LM identification. 
It must additionally run in real-time and its runtime should be easily tunable to allow for adjustments if necessary. 
The fastSLAM 2.0~\protect\cite{Montemerlo2003} algorithm is selected as the particle filter structure inherently allows for computationally efficient independent data associations, compared to multi-hypothesis approaches for EKF-based SLAM~\protect\cite{Montemerlo2010} or graph-based methods. 
Since it does not utilize an optimization step its runtime is lower and more predictable~\protect\cite{Thrun:2005:PR:1121596}, and its accuracy is higher than that of its predecessor (see figure~\ref{fig:slam_comparison} in section~\ref{subsec:slam_results}).
Figure~\ref{fig:slam_arch} illustrates the SLAM architecture. 
The LiDAR and camera pipelines are treated independently, providing observations at different frequencies, with different delays, and different uncertainty models. 
The output of the mapping and localization pipeline is a 2D feature-based map, and a pose within this map, both compensated for delay.
\subsubsection{Mapping phase}
\label{sec:slam_mapping}
The map is updated every time new LM observations are received. 
Each observation consists of a position estimate $\mathbf{z}_{k-{\Delta}k}$ in the local vehicle frame at time step $k-{\Delta}k$ and a color estimate \mbox{$\mathbf{c}_{obs,k}=\mathbf{c}_{obs,{k-{\Delta}k}}=[p_{blue}, p_{yellow}, p_{orange}, p_{other}]^T$}. 
The delay is compensated by propagating the observations forward using the motion estimate between time step $k-{\Delta}k$ and $k$.
To update the filters for each particle, a new particle pose is predicted through an odometry motion model without noise~\protect\cite{Thrun:2005:PR:1121596}. 
The LM observations are then associated to already existing LMs in the map using the nearest neighbour method \protect\cite{nieto2003}, where the Mahalanobis distance is used as a measure for the likelihood of correlation and to find the data association $\mathbf{a}_k$ for each particle. 
If the maximum likelihood of an observation correlating to any LM is below a threshold $l$, the observation is assumed to belong to a new LM. 

The pose accuracy is enhanced by an EKF which is iteratively refined by the incorporation of the matched observations. 
After drawing a pose from the generated distribution, the position estimates of the observed LMs are updated in a straight forward manner using standard EKF equations.
\subsubsection{Color Integration}
\label{sec:slam_color}
The LM colors are modelled as a categorical distribution with $K=3$ possible outcomes. 
A color estimate is drawn from the distribution provided by each sensor. 
The respective counter $\alpha_{i}$ is then increased. 
In the color probability distribution of each LM, the probabilities are set to their expectation value~\protect\cite{Minka03bayesianinference}:
\begin{equation}
    p_i = \mathbb{E}[p_i] = \frac{\alpha_{i}}{\sum_{k=1}^{K} \alpha_{k}}
\end{equation}
Depending on the number of measurements received and the type of sensor, a color is drawn from the respective distribution using the maximum-a-posteriori method.
\subsubsection{Particle Weighting}
\label{sec:slam_weighting}
Each particle is weighted according to how well the observations match the already existing map. 
The total weight $w_{k}$ of a particle at time step $k$ is given by:
\begin{equation} 
    w_{k} = w_{k-1} * l^{\nu} * w_{b}^{\kappa} * w_{c}^{\gamma} * \prod_{n \in \mathbf{a}_k} w_{k,n}
\end{equation}
where $\nu$ is the number of new LMs, $\kappa$ the number of LMs that are in sensor range but were not observed (penalized by $w_{b}$ per LM), $\gamma$ the number of LMs whose color did not match (penalized by $w_{c}$ per LM) and $w_{k,n}$ the weights of all matched LMs, which are computed after updating the EKF for each LM. 
The weights of all particles are then normalized. 
The particle weight variance naturally increases over time and therefore resampling is enforced once the effective sample size $N_{eff}$ over the total number of particles drops below a given threshold.
\subsubsection{Failure Detection}
\label{sec:slam_failure}
For each new set S of LM observations, a sensor failure detection step is applied after data association and used to reduce map quality degradation due to irreversible EKF updates. The observation ratio of a LM is defined as the number of times the LM has been detected over the total number of times it was in a sensor's perceptual field of view (FoV). The set S is only accepted if enough observations match with landmarks that have an observation ratio above 80\%, given there are any in the FoV. 
\subsubsection{Post-Processing}
\label{sec:slam_post}
After one driven lap, a track loop is detected when all particles collapse within \SI{4}{\meter} around the start with a standard deviation of less then \SI{0.2}{\meter} and a similar orientation compared to the beginning of the race. 
Subsequently, the boundaries of the track are estimated using the map of the highest-weighted particle. 
The LMs are classified as inside and outside and ordered according to the previously driven line. 
\subsubsection{Localization}
\label{sec:slam_localization}
After the first lap - in case of loop closure - the EKF map update of the fastSLAM 2.0 algorithm is disabled, which essentially turns it into Monte Carlo localization. 
The pose estimate is computed as the mean of all particle poses. 

	\label{subsec:slam}

\section{Results}
\label{sec:results}
	
The approaches proposed in this paper are deployed on the autonomous race car \textit{gotthard driverless}. It is based on \textit{gotthard}, an electric 4WD race car with a full-aerodynamics package built by AMZ\footnote{www.amzracing.ch} in 2016.
Additional sensors and actuators were added to enable fully autonomous operation. The car was tested in five different testing locations in addition to the competitions, and all the data presented in this section has been gathered during these runs.

\subsection{LiDAR Cone and Color Detection}
\label{subsec:lidar-results}

\begin{figure}
  \centering
  \begin{subfigure}[b]{0.8\columnwidth}
  	\centering
   	\includegraphics[width=\linewidth]{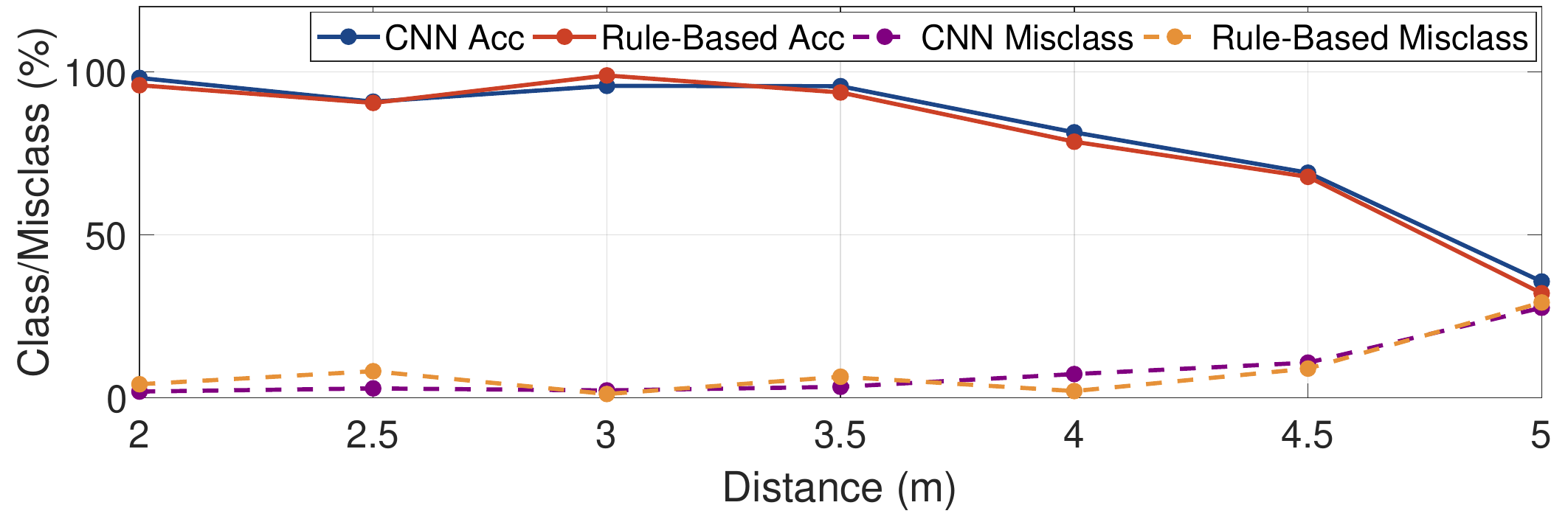}
   	\setlength{\abovecaptionskip}{-11pt}
    \setlength{\belowcaptionskip}{0pt}
    \caption{}
    \label{fig:lidar-color-model-comparison-graph}
  \end{subfigure}
  \begin{subfigure}[b]{0.8\columnwidth}
  	\centering
   	\includegraphics[width=\linewidth]{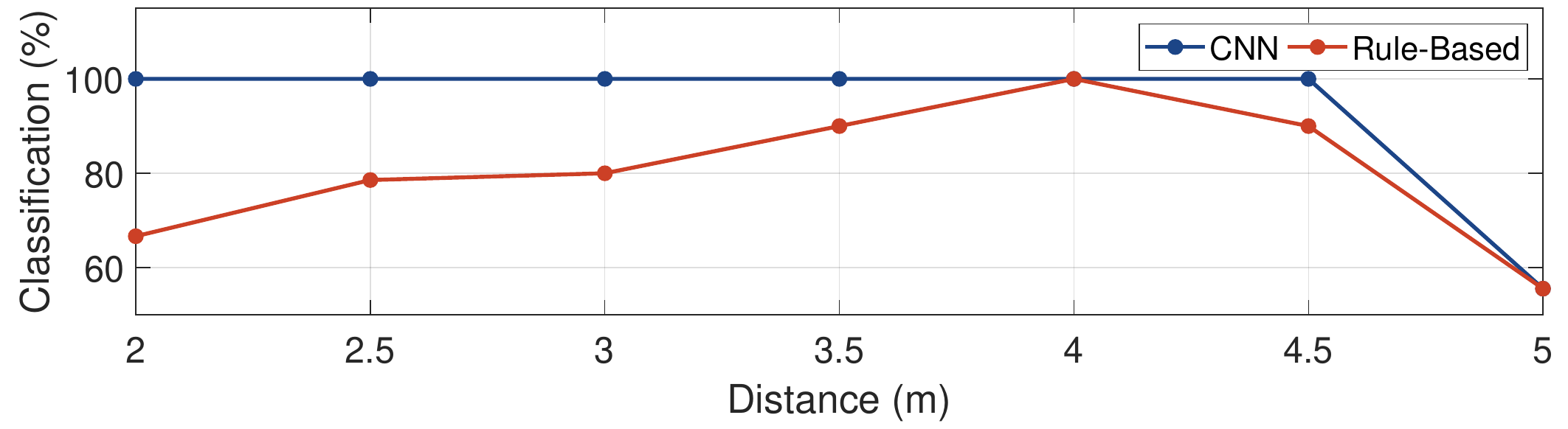}
    \setlength{\abovecaptionskip}{-11pt}
    \setlength{\belowcaptionskip}{-0pt}
    \caption{}
    \label{fig:lidar-color-model-comparison-new-data-graph}
  \end{subfigure}
  \setlength{\abovecaptionskip}{1pt}
  \setlength{\belowcaptionskip}{-5pt}
  \caption{Classification performance of the CNN and the rule-based approach when (a) using the same cones as in training and (b) different ones.}
\end{figure}

Figures \ref{fig:lidar-color-model-comparison-graph} and \ref{fig:lidar-color-model-comparison-new-data-graph} compare the performance of the CNN and the rule-based approach on datasets not used for training. 
For cones of the same type as the ones presented during training, both approaches provide similar results, giving an accuracy of around $96\%$ for the ones close-by. 
However, the difference between the two arises if different cone types are used during testing. 
In the competition in Germany, the cones have an \emph{FSG} sticker that results in different point cloud intensities. 
The rule-based approach shows a larger number of misclassifications, whereas the CNN is hardly affected by it.
This supports our initial hypothesis that the CNN is more reliable and generalizes better as compared to its rule-based counterpart.

These figures also show the significant drop in the accuracy when the distance is around \SI{5}{\meter}, justifying the decision to cap color estimation at this distance. Most of the cones at such distances are labelled \textit{unknown} which results in reduced classification accuracy, but not increased misclassification because a blue cone is not labelled yellow and vice-versa. This reduces the number of false color estimates, thus ensuring the robustness of the system.

\subsection{Visual Cone Detection and Stereo Pose Estimation}
\label{subsec:stereo-results}

\begin{figure}
	\centering
  \begin{subfigure}[t]{0.51\columnwidth}
   	\includegraphics[width=\linewidth]{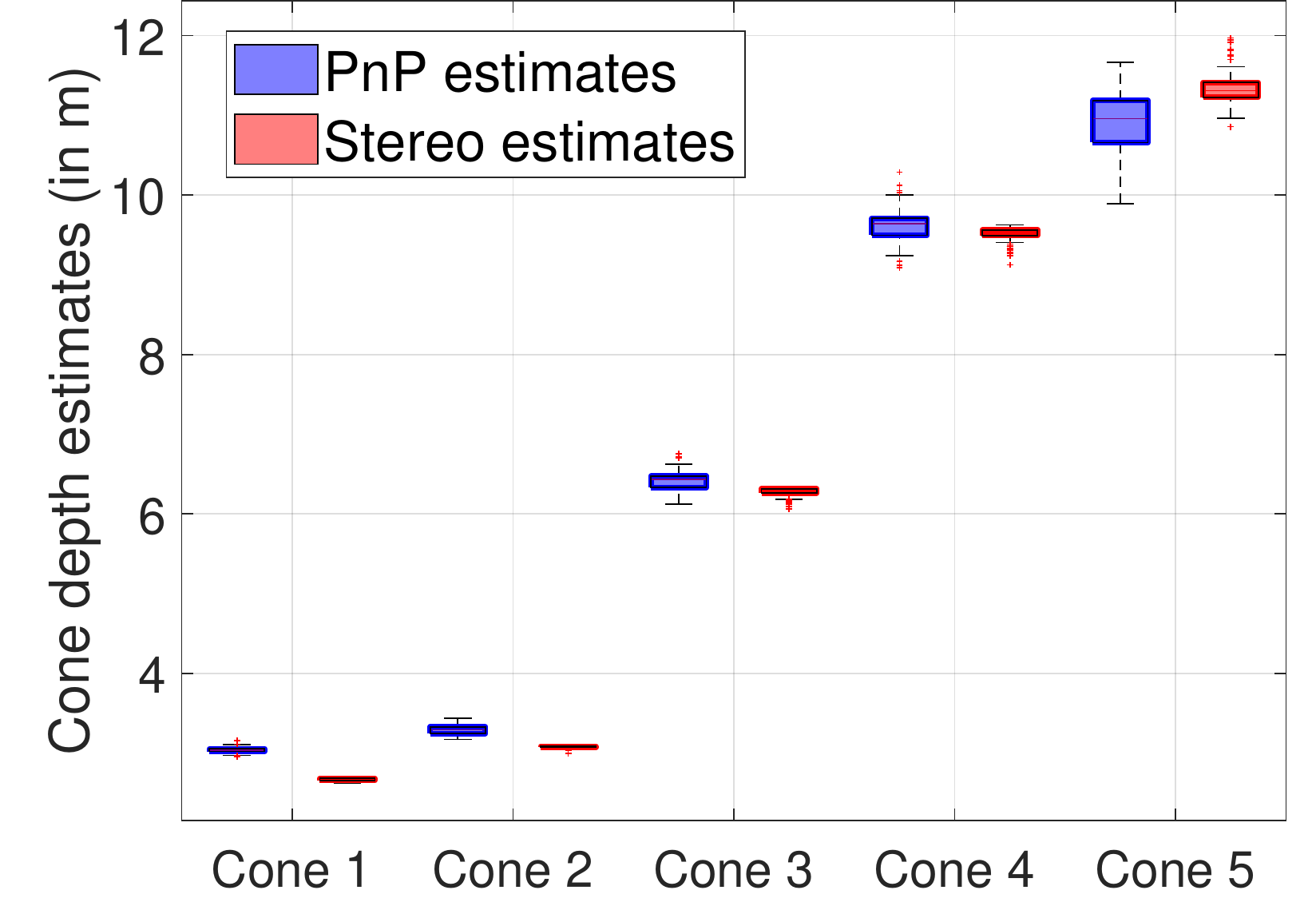}
   	\setlength{\abovecaptionskip}{-10pt}
   	\setlength{\belowcaptionskip}{0pt}
    \caption{}
    \label{fig:vision-stereo-pnp}
  \end{subfigure}
  \begin{subfigure}[t]{0.39\columnwidth}
    \includegraphics[width=\linewidth]{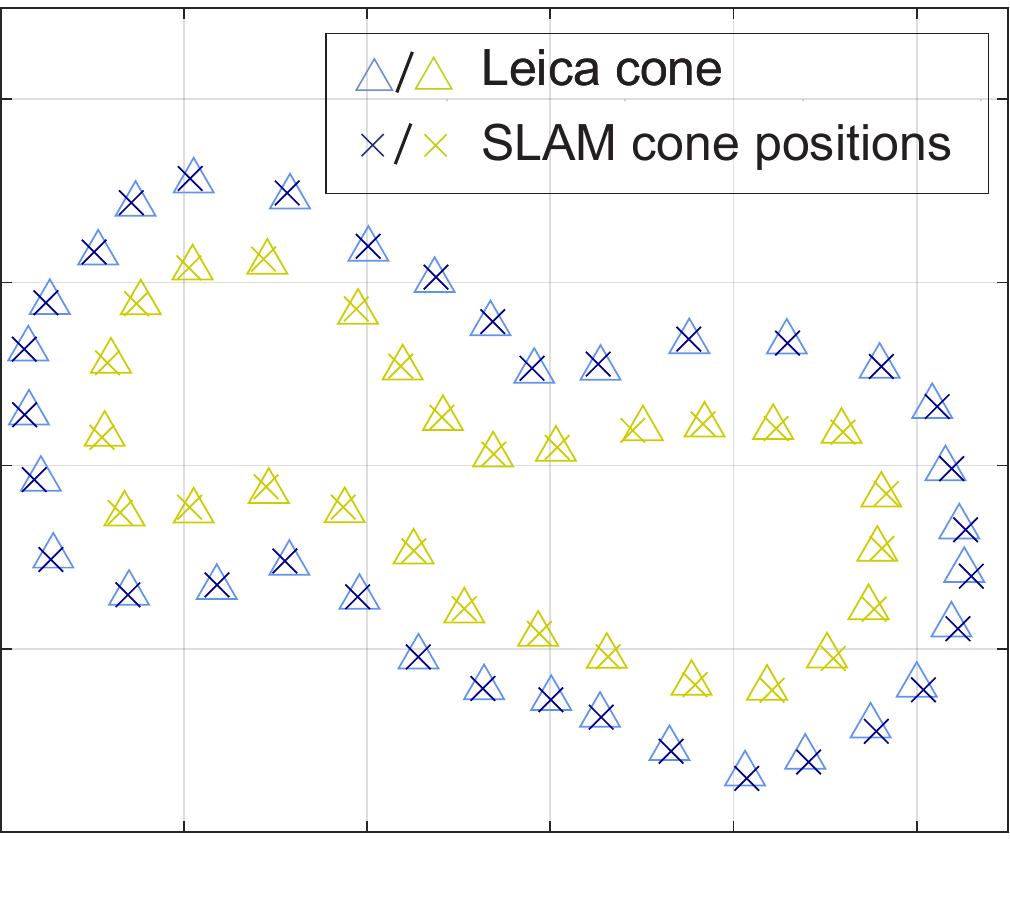}
   	\setlength{\abovecaptionskip}{-10pt}
   	\setlength{\belowcaptionskip}{-20pt}
    \caption{}
    \label{fig:vision-slam}
  \end{subfigure}
  \setlength{\abovecaptionskip}{0pt}
  \setlength{\belowcaptionskip}{-20pt}
  \caption{(a) Box plots of depth estimates of five cones obtained from the left camera's image via PnP and via the stereo approach using triangulation (b) Resulting map computed in real-time using fastSLAM 2.0 implementation in the mapping phase, compared to a ground truth measurement using the Leica Totalstation. The grid size is 10m x 10m.}
\end{figure}
Figure \ref{fig:vision-stereo-pnp} compares the depth estimates of cones obtained from the PnP algorithm to those obtained through triangulation. Multiple measurements of the same scene are taken to illustrate the variance of the estimates. It can be observed that the variance is reduced significantly by using triangulation, especially for cones that are far away. However, due to the fact that disparity decreases with distance, the position estimates' accuracy drops at larger distances. Hence, the maximum depth estimate provided by the stereo setup is limited to \SI{10}{\metre}. 

Figure \ref{fig:vision-slam} shows a map generated by SLAM for a \SI{100}{\meter} long track using only the estimates from the stereo camera. A RMS landmark error of \SI{0.25}{\meter} is achieved, which is close to the accuracy achieved when using LiDAR estimates and enough to finish the race in case of a LiDAR failure.

\subsection{Velocity Estimation}
\label{subsec:VE_results}
	
\begin{figure}
	\centering
  \begin{subfigure}[b]{\columnwidth}
  	\centering
   	\includegraphics[width=0.85\linewidth]{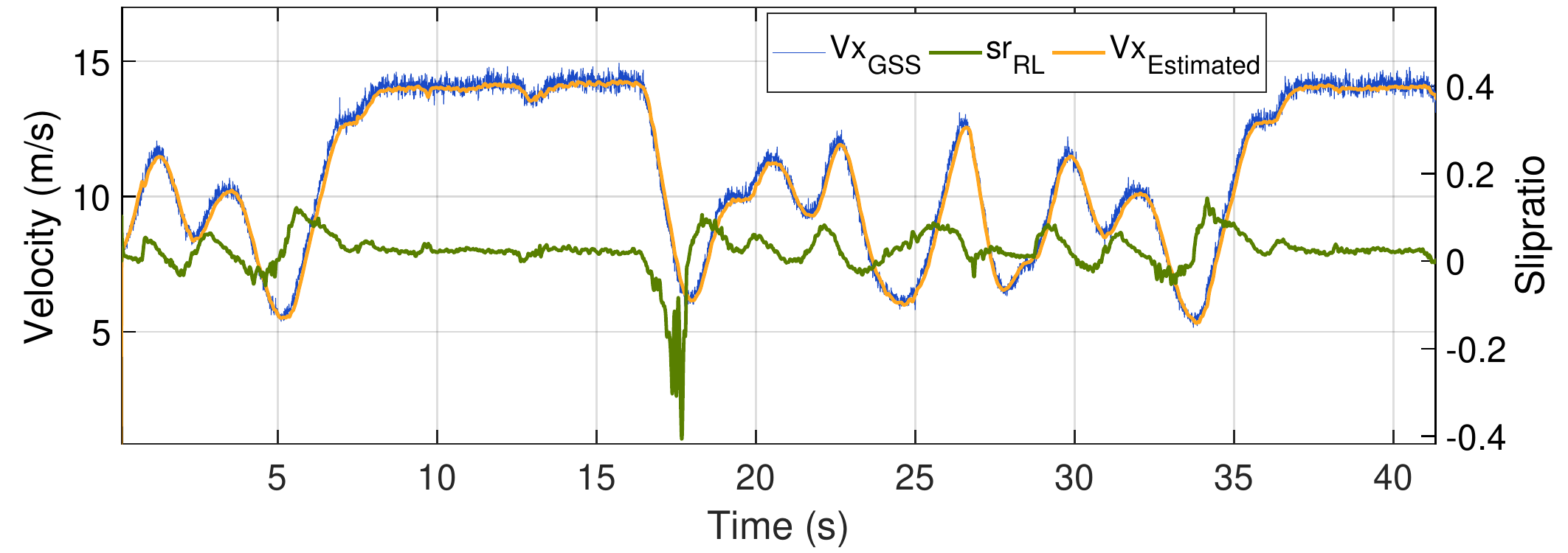}
   	\setlength{\abovecaptionskip}{0pt}
   	\setlength{\belowcaptionskip}{0pt}
    \caption{}
    \label{Fig: VE_long}
  \end{subfigure}
  \begin{subfigure}[b]{\columnwidth}
  	\centering
  	\includegraphics[width=0.85\linewidth]{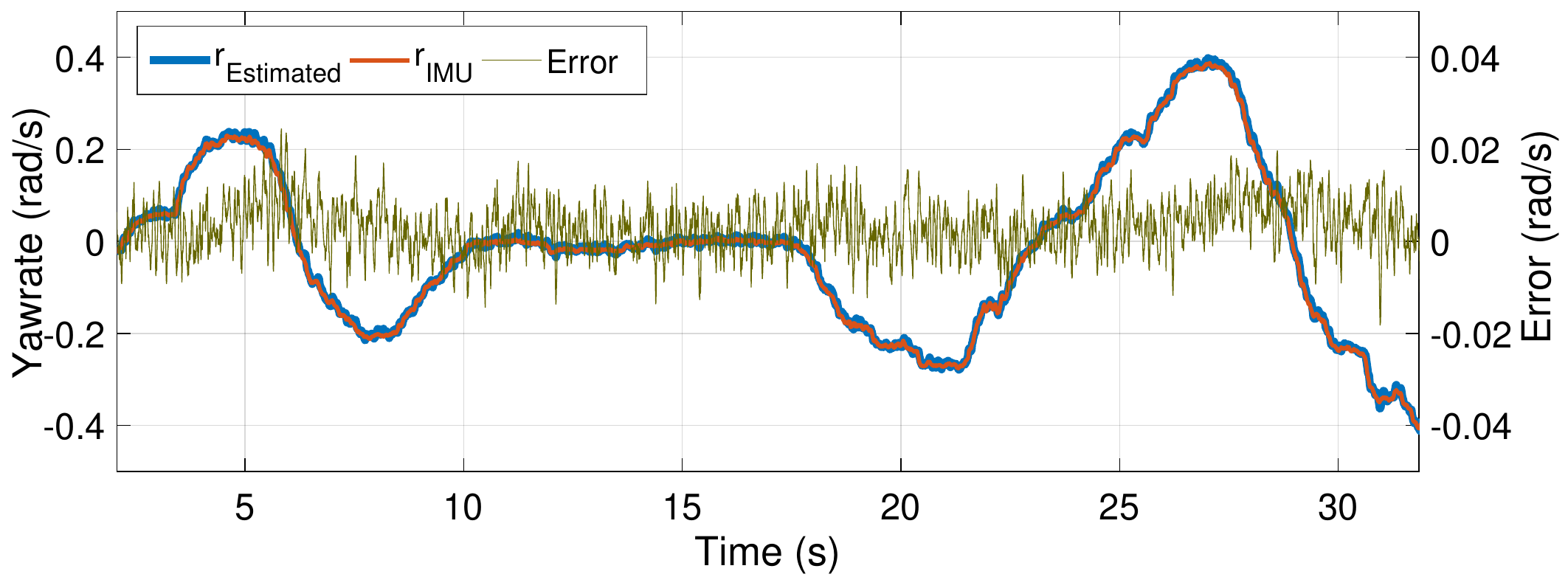}
   	\setlength{\abovecaptionskip}{0pt}
   	\setlength{\belowcaptionskip}{0pt}
    \caption{}
    \label{Fig: VE_yaw}
  \end{subfigure}
  \setlength{\abovecaptionskip}{-13pt}
  \setlength{\belowcaptionskip}{-10pt}
  \caption{(a) Estimated longitudinal velocity as compared to the GSS and (b) yaw rate estimates compared to the IMU.}
\end{figure}

The redundancy of the velocity estimator is analyzed by simulating sensor failures and comparing velocities to ground truth (GT) information based on the GSS data. 
Figure~\ref{Fig: VE_long} shows the estimated velocity without the GSS compared to GT. 
The RMSE is \SI{0.14}{\meter\per\second}. It can be seen that the velocity estimate is accurate even when the wheels slip a lot. 
The distance between the position of the car obtained by integrating the estimated velocity and the GPS position is less than \SI{1.5}{\meter} over a \SI{310}{\meter} long track, which results in a drift of less than 0.5~\%.

Figure \ref{Fig: VE_yaw} compares the estimated yaw rate without IMU to that of the IMU, wherein it is evident that the predicted yaw rate almost converges to the true yaw rate. This implies that the car model is reliable and velocities can be accurately estimated even in the absence of the IMU.

\begin{figure}
	\centering
   	\includegraphics[width=0.9\linewidth]{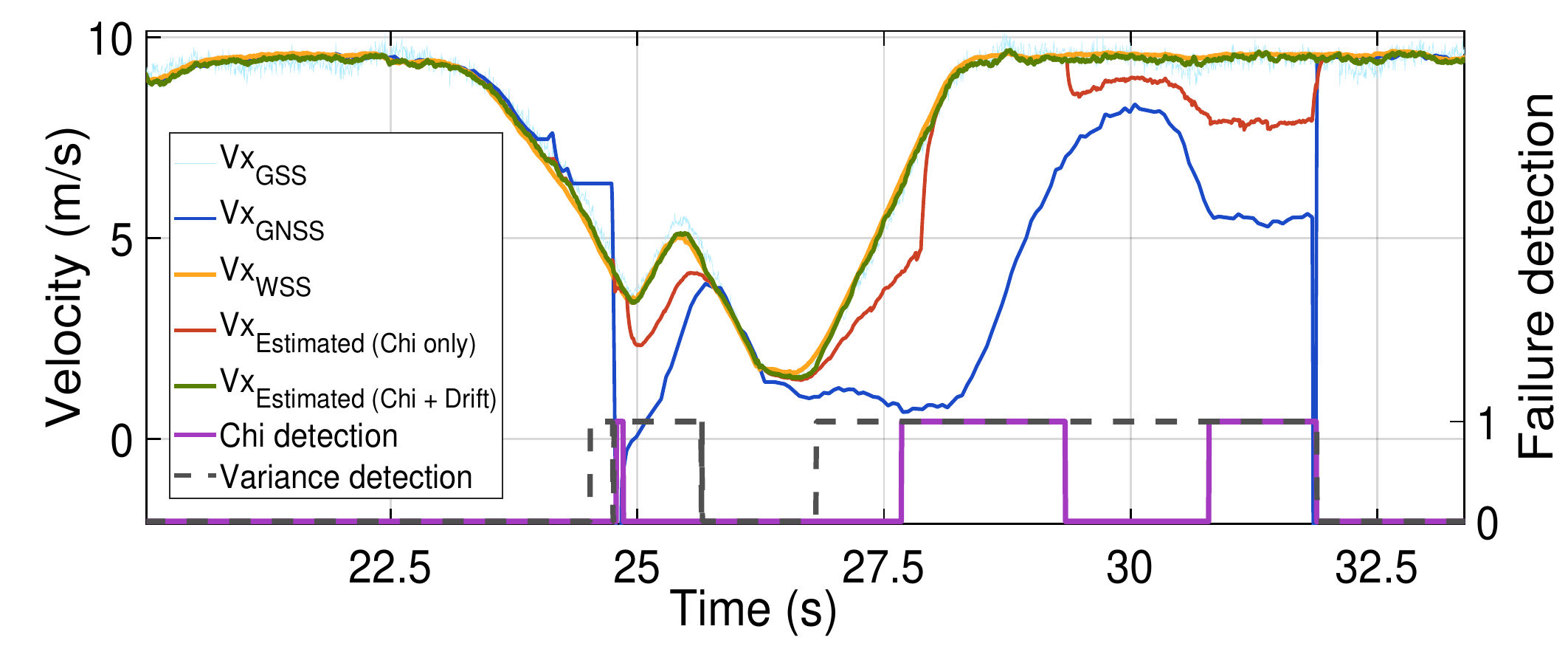}
   	\setlength{\abovecaptionskip}{1pt}
   	\setlength{\belowcaptionskip}{-20pt}
  	\caption{Failure of the GNSS (cyan) is detected by the drift detection (grey dashed). ${Vx_{Estimated (chi \; only)}}$ (red) is the estimated velocity using the chi test (violet) which shows that the chi test without drift detection is unable to detect the full sensor failure.}
    \label{fig:VE_drift_detection}
\end{figure}

Figure \ref{fig:VE_drift_detection} shows the response of the filter to sensor failures. It can be observed that the chi-square based failure detection is able to reject the signal only when the failure is short-lived, whereas the drift failure detection is able to also discard continuous sensor failures. Using both techniques in conjunction ensures removal of most of the sensor failures.

\subsection{Localization and Mapping}
	
\begin{figure}
	\centering
   	\includegraphics[width=0.9\linewidth]{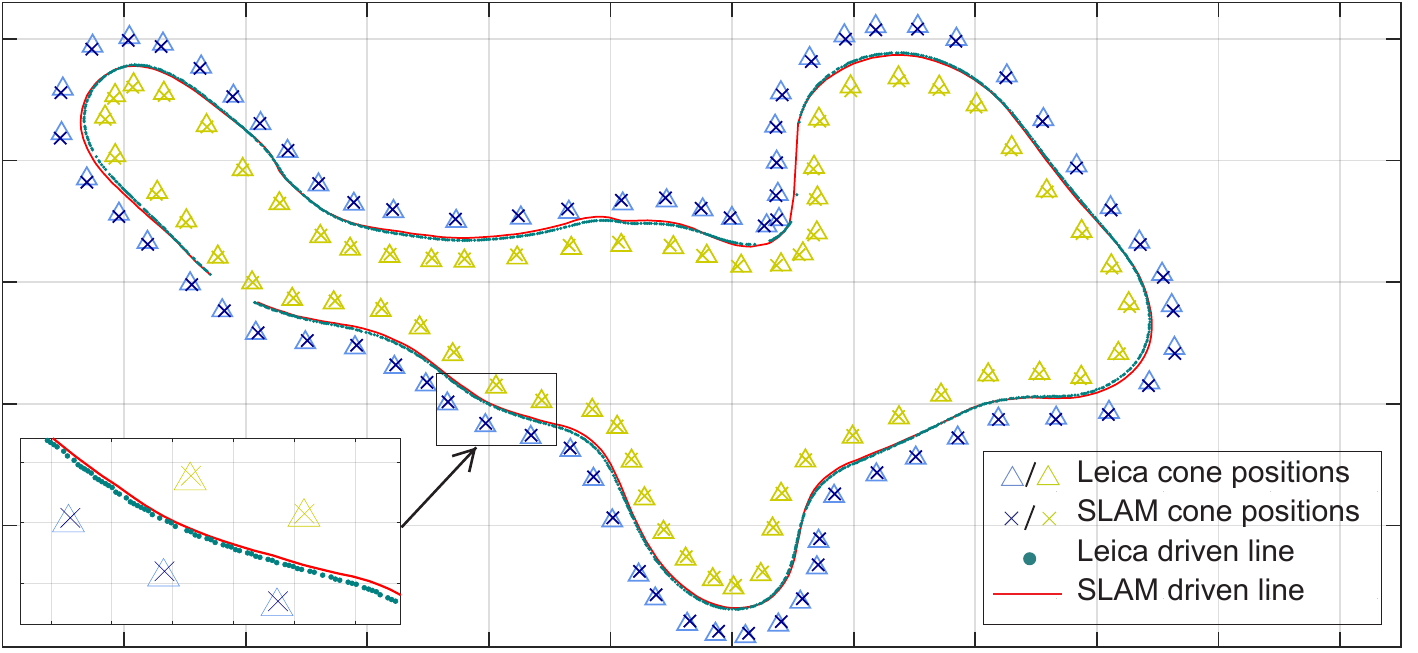}
   	\setlength{\abovecaptionskip}{3pt}
   	\setlength{\belowcaptionskip}{-5pt}
  	\caption{Resulting map and trajectory computed in real-time during the mapping phase, compared to ground truth measurements from Leica Totalstation. The grid size is $\SI{10}{\meter} \times \SI{10}{\meter}$ for the whole map and $\SI{2}{\meter} \times \SI{2}{\meter}$ for the zoomed in image.}
    \label{fig:slam_leica_map}
\end{figure}

\begin{figure}
   \centering
   \includegraphics[width=0.9\linewidth]{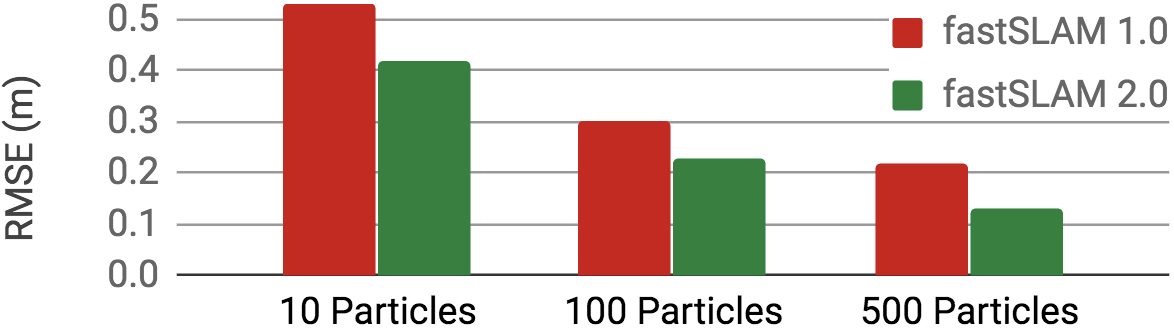}
   \setlength{\abovecaptionskip}{3pt}
   \setlength{\belowcaptionskip}{-20pt}
   \caption{RMS landmark error comparison of fastSLAM 1.0 vs fastSLAM 2.0 for different number of particles against ground truth from a Leica Totalstation.}
   \label{fig:slam_comparison}
\end{figure}

The accuracy of the SLAM algorithm is evaluated by comparison against ground truth (GT) measurements from a Leica Totalstation.
Figure \ref{fig:slam_leica_map} shows the result for a \SI{230}{\meter} long track using LiDAR color detection and SLAM only. An RMSE of \SI{0.2}{\meter} is obtained for both landmarks and the driven path. All cones' colors are correctly estimated.
A performance comparison of fastSLAM 1.0 vs 2.0 is shown in figure \ref{fig:slam_comparison}. The results show that the RMSE for a given amount of particles is significantly smaller for fastSLAM 2.0., which implies a higher map accuracy. 

\label{subsec:slam_results}

\section{Conclusion}
\label{sec:conclusion}
	
This paper presented the approaches developed to ensure reliable operation of an autonomous race car by introducing redundancy into the perception and state estimation pipelines. 
It has been shown that accurate color estimates can be obtained from LiDAR using the intensity signature of the point clouds, and accurate positions can be estimated from cameras using prior knowledge of objects.
Additionally, we have demonstrated accurate velocity estimation during high wheel slip and under single-sensor failure which could be correctly identified by the failure detection module.
In future work, it would be interesting to investigate whether the performance of the failure detection module can be improved by performing post-state analysis.
Finally, the fastSLAM 2.0 algorithm has been adapted to map and localize in real-time using the output of either one or both perception systems.
Extensive testing shows that the algorithms generalize well to unseen environments, even under sensor failure, thus paving the way towards autonomous race vehicles that can drive close to the limits of handling.

\section*{ACKNOWLEDGMENT}
The authors thank the AMZ Driverless team for their sustained hard-work and passion, as well as the sponsors for their financial and technical support. 
We also express our gratitude to Marc Pollefeys, Andrea Cohen, and Ian Cherabier (CVG Group, ETH Z\"urich) for their support throughout the project.

\newrefcontext[sorting=none]
\printbibliography
\end{document}